# Prediction, Expectation, and Surprise: Methods, Designs, and Study of a Deployed Traffic Forecasting Service

**Eric Horvitz**
Microsoft Research
Redmond, Washington
*horvitz@microsoft.com*

**Johnson Apacible**
Microsoft Research
Redmond, Washington
*johnsona@microsoft.com*

**Raman Sarin**
Microsoft Research
Redmond, Washington
*ramans@microsoft.com*

**Lin Liao**
University of Washington
Seattle, Washington
*liaolin@cs.washington.edu*

## Abstract

We present research on developing models that forecast traffic flow and congestion in the Greater Seattle area. The research has led to the deployment of a service named *JamBayes*, that is being actively used by over 2,500 users via smartphones and desktop versions of the system. We review the modeling effort and describe experiments probing the predictive accuracy of the models. Finally, we present research on building models that can identify current and future surprises, via efforts on modeling and forecasting unexpected situations.

## 1 Introduction

We describe our work to develop and field a traffic forecasting service that monitors traffic patterns and that relays predictions about traffic forthcoming congestions and flows to users in mobile settings. The efforts have led to a prototype traffic service that is now relied upon daily by over 2,500 people. The predictive models take into consideration information about the status and dynamics of traffic in the Greater Seattle area. Other evidence considered in learning and prediction includes incident reports issued by the Washington Department of Transportation, information about the occurrence of major sporting events, weather reports, time of day, and calendar information.

We discuss the modeling effort and review studies of the predictive accuracy of the learned models. After discussing details about the base-level predictions of the service, we discuss how we mesh inferences with visual representations that can communicate probabilistic information with a quick glance. We also examine route-based alerting, based on a specification of user interests and preferences. We then turn to the challenge of constructing models that have the ability to predict when users may be surprised about sensed traffic events. Finally, we review our approach to learning models that can predict *future surprises* about traffic congestion and flow. This effort provides a window on possibilities for employing surprise forecasting in other realms. The models, visualization, and related components highlight the challenges and opportunities for leveraging machine learning and reasoning in a mobile application that can provide ongoing value to a large group of people.

## 2 JamBayes: A Traffic Forecasting Service

We have been intrigued by the challenges and opportunities for providing information and inferences to people in mobile settings. In one approach, we can embed reasoning machinery or compiled policies in small, portable devices that perform local, real-time sensing. For example, in the Bayesphone project (Horvitz, Koch, Sarin, *et al.,* 2005), decision-theoretic policies, based on models learned on desktop computing systems, are downloaded onto smartphones. The devices are endowed with the ability to decide when to engage users with questions about context and to take call-handling actions based on expected utility. In another approach, policies, recommendations, and information can be streamed from server-based learning and reasoning systems to portable devices, based on information from the devices and other sources. We focus on the latter approach in this paper, a methodology we refer to as *streaming intelligence*.

We constructed and fielded a web service that provides users with the current status and predictions about the future of traffic flow at key hotspots within the Seattle highway system. The host application, named Smartphlow, provides users with a visualization of current traffic flows in Seattle and surrounding areas, by displaying color coded segments on major arteries to relay the speeds and densities of cars. The basic display of traffic status relays information reported by a network of sensors operated by the Washington Department of Transportation (WDOT). Where there is a smooth, fast flow of traffic, cells reflecting sensed regions of the highway system, are colored green. As roads become more congested the color coding of segments goes from green to yellow, to red, to black, indicating that traffic flow has diminished to a crawl.

Beyond visualizing the current state of affairs on the key roads of Seattle and providing an easy navigation model for translating and zooming on a map of Seattle (Robbins, Cutrell, Sarin, *et al*., 2004), Smartphlow provides predictions about the future status of regions of the traffic system. We shall focus on the predictive component of Smartphlow, named JamBayes, a service that continues to provide forecasts to users about the likely times until congested bottlenecks will likely free up, and when key locations of the highway system that are currently open will likely become jammed.

We made Smartphlow widely available to Microsoft employees nearly two years ago, and it has grown through word of mouth to become a popular application. We are logging unique hits to JamBayes servers and note that the service is now relied upon daily by over 2,500 users. We shall review key aspects of Smartphlow and the JamBayes prediction service in the next several sections.

## 2.1 Identifying Key Bottlenecks

Numerous analyses in traffic engineering have explored the microstructure of flows at cells, using such methods as queue-theoretic models (*e.g.,* Vandaele, Van Woensel, and Verbruggen, 2000). Rather than model flows explicitly, we sought to abstract the problem of traffic prediction in the Greater Seattle area to a consideration of probabilistic dependencies among a set of random variables, representing properties of key congestion "hotspots" and contextual observations. Such hotspots are well-known by Seattle commuters, as they are often the source of frustrating delays, even when traveling short distances on the arterial system.

We believe that representing and reasoning about the status of key bottleneck regions can serve as an organizing principle for building predictive traffic systems for cities throughout the world. Our research on adapting our traffic prediction methodology to other cities in the country and world has highlighted the commonality of sets of well-known hotspots. Such trouble spots often acquire nicknames, like "the S-curves," that are bandied about in conversations and reports by traffic reporters during morning and evening commuting.

The identification of a set of hotspots, which we refer to as traffic *bottlenecks*, for a traffic system enables us to focus the attention of modeling and alerting to a set of events and states that people care deeply about. Such a framing helps to reduce the parameter space of the learning and inference effort as well as provide a representation of traffic problems that is easy to communicate with users.

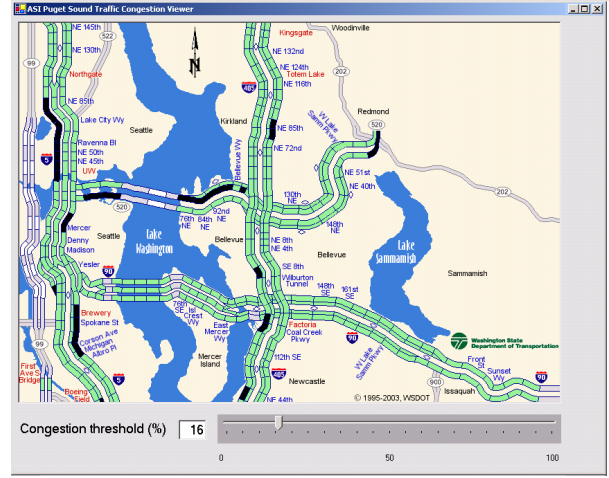

Figure 1: Bottleneck identification tool used to frame the learning problem for modeling Seattle traffic. Key trouble spots in the traffic system are highlighted via setting a threshold on percent time cells are congested.

To probe Seattle's bottlenecks, we developed an interactive tool that analyzes a large database of system-wide traffic flow data collected over many months. The tool allows the moving of a slider to set a threshold amount of "percent of time of day cell is congested," for all monitored cells. When moving the slider, the cells of the highway system that are congested for at least the time represented by the current setting of the threshold are colored black. As the application user moves the threshold from 100% to 0% of the total time of the day, the most frequently congested spots, initially appearing as small regions of black segments, extend and meld together until the whole traffic system is marked. A screenshot from the bottleneck analysis tool is displayed in Figure 1.

Using the tool, we identified 22 key bottlenecks in the Seattle traffic system. Given the formulation of the hotspots, we set out to learn statistical models that could provide inferences with users about the expected time until a bottleneck, that was currently experiencing a traffic jam, would flow smoothly again, and the time until bottlenecks, which were currently flowing smoothly, would likely become jammed.

## 2.2 Collecting a Case Library and Learning Models

We began collecting traffic data and related information nearly two years ago by monitoring and storing as cases the status of all sensed traffic cells in the Seattle system. We also collected terse natural language "incident reports" that are emitted from time to time by operators at the WDOT. Incident reports include mention of a variety of traffic problems, such as accidents of different scales, including major incidents requiring emergency vehicles that span multiple lanes. A sample incident report is displayed in Figure 2.

Beyond traffic data, we have collected contextual data that promised to be informative about traffic flows. These observations include the time of day, day of week, whether a holiday is in progress, whether school is in session, and the scheduling of large-scale events in the city, such as major sporting events like Mariners, Sonics, Huskies, and Seahawks games. We also included in the case library synchronized information about Seattle weather, including the status of precipitation, visibility, temperature, and sunshine. We have continued to harvest in an automated manner weather and event information from Web sources.

```
Operator ID: Nick
Heading: INCIDENT
Message:INCIDENT INFORMATION
Cleared 1637: I-405 SB
JS I-90 ACC BLK RL CCTV
1623 – WSP, FIR ON SCENE
```

Figure 2: A typical incident report, mentioning police and fire vehicles within a particular bottleneck region.

For each identified bottleneck region, we created sets of random variables that represent static and temporal abstractions of the status of one or more sensed cells within that bottleneck. Such random variables include the number of cells that are showing blockage or slowing, the time since any cell became blocked within the bottleneck region after a threshold period of time when all cells showed open flow, and the maximum number of adjacent blocked cells.

Given the rich case library, we experimented with several machine learning formulations and methods. As part of the iterative process of building and refining models, we created new distinctions about traffic, including several summarizing statistics capturing recent dynamics in flow, such as the rate and density of changes in velocities in cells within and near bottlenecks over time. Figure 3 shows an overview of the properties of cases and overall process of generating a graphical model for predicting traffic jams and flows.

In the current fielded version of JamBayes, we employ Bayesian structure search, using tools developed by Chickering, *et al.* (Chickering, Heckerman, and Meek, 1997) to construct a Bayesian network. The method provides a graphical view onto the model, enabling us to visualize multiple variables and influences. Given a training dataset, the method performs heuristic search over a space of dependency models using a Bayesian scoring criterion to guide the search. Details on the heuristic search and model scoring can be found in (Chickering, Heckerman, and Meek, 1997). We employed a mix of discrete and continuous variables in the models, using continuous variables to represent times of different events interest. For each discrete variable, the method creates a tree containing a multinomial distribution at each leaf. For each continuous variable, the method constructs a tree in which leaves contain a *binary-Gaussian* distribution; each leaf contains a binomial distribution that represents the probability of the value being present or absent, and a Gaussian distribution over the values that are present. The Bayesian score used for each binary-Gaussian distribution is the sum of the score for the binomial, a special case of the multinomial, and the score for a Gaussian defined over the values that are present in the training data. For details of the Gaussian score, see Heckerman and Geiger (1995).

Rather than build separate models for each bottleneck, we construct a large model for all bottlenecks. The comprehensive model captures the rich interdependencies among multiple bottlenecks and other variables, allowing the system to learn about dependencies and temporal relationships among the flows and congestions at bottlenecks.

Figure 4 highlights, as an example, the influences within a learned graphical model of the *weather* variable, representing mutually exclusive sets of states about weather conditions, on other variables in the model, including variables capturing the amount of time until bottlenecks will become jammed if they are currently open, and the time until bottlenecks will melt into flows should they be currently bottlenecked. We shall explore the performance of the model after we briefly review how predictions are relayed to users.

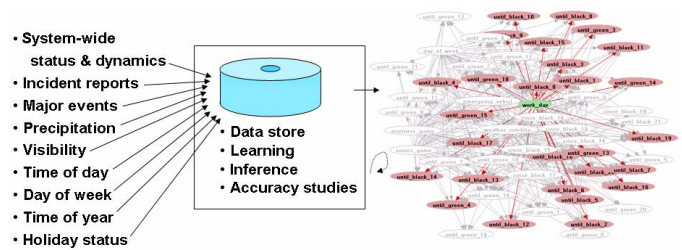

Figure 3: Overview of learning predictive models for traffic flows at potential bottlenecks. Cases include data from the highway system, incident reports, major events (*e.g.,* sporting events), weather, time of day, day of week, and holiday status.

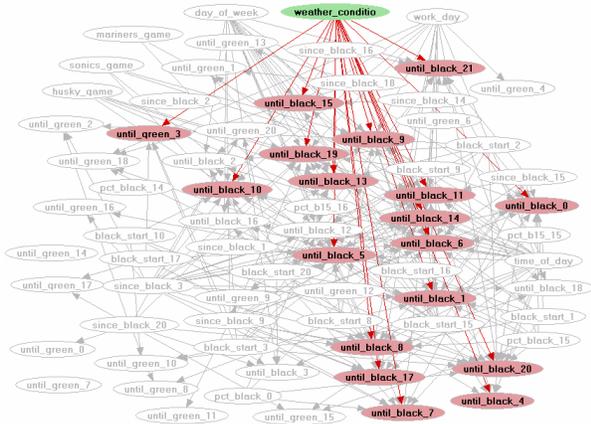

Figure 4: Predictive model highlighting influence of weather conditions on the time for clearing and jamming for multiple portions of the traffic system.

### 2.3 Visualization for Relaying Inferences

Mobile users are typically engaged in a great diversity of primary tasks, such as driving a vehicle, when they may wish to access traffic status and predictions. Thus, it is critical to design lightweight navigation methods and visualizations for relaying predictions that can be glanced at quickly. Users can access different regions of the highway system in Smartphlow by depressing the 1-9 dialing keys of the phone. The keys map isomophically to regions of the map. Users can also toggle between two levels of zoom depressing the smartphone joystick. By depressing the 0 button, users are taken on a dynamically constructed flyover that dips down and dwells briefly on current troublespots, before returning to the starting point. Figure 5 shows the display we employ on smartphones for communicating JamBayes inferences. If a bottleneck is currently congested, a "clock" graphic appears next to the congested segment. The clock is filled with red proportional to the maximum likelihood time that the congestion will persist before it becomes a flowing thoroughfare. If a bottleneck is to become congested in an hour or less, a clock filled with green appears, filled to represent the time expected until a jam appears.

As highlighted in the inset in Figure 5, we also display the confidence in the prediction provided by the model with tick marks around the maximum likelihood, showing the computed standard deviation around the inference.

### 2.4 Study of Prediction Quality

We have continued to perform studies of the accuracy of the JamBayes service. Table 1 shows results from a recent evaluation of the time until congested bottlenecks clear, and the time until clear bottlenecks jam, for each bottleneck, based on 15 months of data. We trained the model with seventy-five percent of the data and tested the model on the remaining twenty-five percent of the cases, segmented sequentially. In the study, we considered an outcome a success if the prediction was within 15 minutes of the predicted time. Because the system relays to the user specific times up until 1 hour, and then indicates a prediction as being one hour or more with a filled clock icon, we consider a prediction of an hour or more as being a correct classification if the time until clearing turns out to be more than one hour. Classification accuracies fall off if we attempt to reason about the details of longer durations, as, for example, traffic can be open for many hours before a bottleneck shows up, posing a more difficult prediction task.

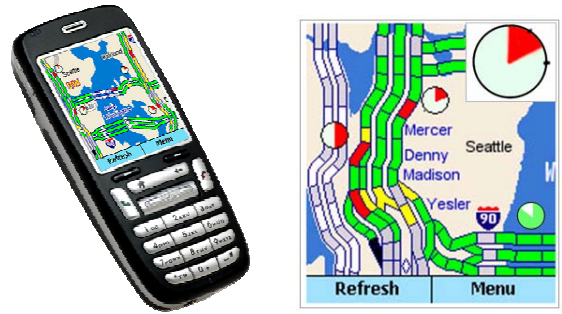

Figure 5: Smartphone application. Clock graphics are used to display time until a congested bottleneck will clear (red) and the time until a bottleneck that is currently open will become congested (green). Inset: close up of prediction, showing confidence as represented by the standard deviation.

Table 1: Accuracy of predictions for time until jams will clear and will form (15 minute tolerance).

| Bottleneck | Accuracy (Clear, Jam) | Bottleneck | Accuracy (Clear, Jam) |
|---|---|---|---|
| 0 | 0.83, 0.92 | 11 | 0.76, 0.93 |
| 1 | 0.75, 0.93 | 12 | 0.65, 0.98 |
| 2 | 0.78, 0.91 | 13 | 0.70, 0.95 |
| 3 | 0.83, 0.90 | 14 | 0.83, 0.95 |
| 4 | 0.87, 0.96 | 15 | 0.80, 0.84 |
| 5 | 0.73, 0.94 | 16 | 0.73, 0.86 |
| 6 | 0.65, 0.95 | 17 | 0.78, 0.92 |
| 7 | 0.84, 0.97 | 18 | 0.76, 0.86 |
| 8 | 0.85, 0.92 | 19 | 0.68. 0.94 |
| 9 | 0.81, 0.96 | 20 | 0.82, 0.96 |
| 10 | 0.71, 0.91 | 21 | 0.86, 0.94 |

### 2.5 Learning Models of Competency

In addition to studying the accuracy of the learned model for predicting the times until bottlenecks will clear or will form, we also studied the value of learning separate *reliability models* for each bottleneck. The reliability models predict whether a base-level prediction will fail to be accurate. This effort comes in the spirit of our work to make systems exploiting learning and reasoning more usable by providing users with context-sensitive competency information. Our intuition was that we would be able to predict the likelihood that the base-level model would provide accurate predictions with greater accuracy than the more difficult, detailed base-level predictions.

To build reliability models for a bottleneck, we execute the learned base model on a test set, and label predictions as falling within a predefined time tolerance for a bottleneck event. We then use this new case library, tagged by success and failure, to build a bottleneck-specific reliability model. In use, the reliability model predicts whether or not the base level model's output will be within a predefined tolerance given all the observations available to the system. We found that the reliability models were typically simpler than the base-level models. The studies of reliability models demonstrated that, for most bottlenecks, we could construct models that could predict when models would fail to provide predictions within a 15 minute tolerance of an outcome with greater accuracy than the marginal performance associated with the classification accuracies of the base-level model. Sample models for the reliability of predictions of the base-level model on the duration of a jam at bottleneck 1 and 11 are displayed in Figure 6. The learned models show that the reliabilities are influenced by a small number of variables, representing such variables as the durations and extent (percent cells showing blockage) of jams at relevant bottlenecks.

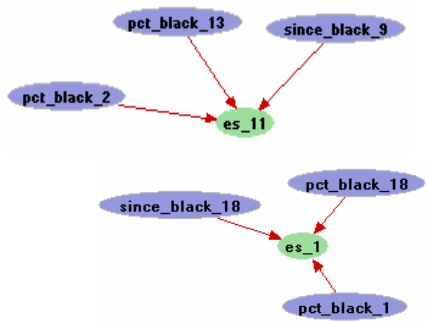

Figure 6: Sample learned models of context-sensitive accuracy of the reliability of predictions of the base model for bottlenecks 1 and 11.

We learned models for reliability for each bottleneck and harnessed these models on the JamBayes server to annotate predictions provided by the base-level models. If the overall accuracy of the reliability models had been found to be high, and the reliability model predicts in real time that, in the current situation, the accuracy of the base-level models is lower than a reliability threshold, the system overlays a question mark within the icon as displayed in Figure 7. This provides users with feedback about when the prediction may be inaccurate.

Informal feedback on the use of competency annotation has been positive. The use of certifiably accurate reliability models that perform a task simpler than a more complex base-level challenge underscores a valuable direction in building systems that share important, but intermittently inaccurate inferences with people. We believe that providing systems with accurate models of their context-sensitive competency, and joining such reasoning with a means of communicating such inferred competency, *e.g.*, with simple visual representations, will enable users to understand when they can trust systems that may fail intermittently.

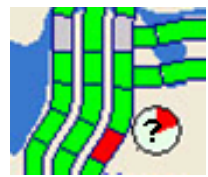

Figure 7: Use of a question mark annotation to relay to the user that prediction may be unreliable in the situation at hand, based on inference with a context-sensitive reliability model for a bottleneck.

To be sure, a critical goal is to make base-level models better rather than build reliability models for underlying "black box" learning methods. The accuracy of the learned reliability models suggests that we can raise the accuracy of models by using boosting methods that learn how to handle cases that an initial model had been failing on. Preliminary studies with boosting suggest that employing a staged mixture model approach to boosting (Meek, Thiesson, and Heckerman, 2002) can be effective for increasing the accuracy of the base models. We plan to report on the value of boosting for traffic predictions in a future paper.

### 2.6 Route-Centric Alerting

Beyond direct inspection of the visualizations of the status of traffic, and of the predictions of the JamBayes service, we also provide a means for users to set up time-dependent route-based alerting. The approach allows users to weave together sets of bottlenecks in considered by the system into paths they care about during morning and evening commuting.

We provide JamBayes users with a desktop tool called Deskflow, which provides JamBayes inferences in desktop settings, and also allows for configuration of alerting on mobile devices. A screenshot from Deskflow is displayed in Figure 8. This mode of the program enables users to define routes of interest for their morning and evening commutes by defining routes as sets of bottlenecks for different segments of time.

Active periodic refreshing of the device with fresh information can be limited to these specific periods of time so as to conserve battery life and data costs.

After specifying routes, users can also set up route-centric alerting policies. When alerting policies are set up, users receive audio and vibratory alerts when specific criteria are met, based on the time of day. Figure 9 shows the configuration for alert configuration. As examples of policies, users can set the system to register an alert on a smartphone, or send an SMS message to a standard cell phone, when their specified route becomes congested or becomes uncongested—or when their route will likely become congested or become free flowing within a selected period of time. As an example, for the configuration represented by the screenshot in Figure 9, a user has instructed the system to send an alert, during specified morning and evening hours, when their route will likely become congested in 30 minutes, when their route becomes clear after it has been blocked, and when the route will likely become clear in 20 minutes, *e.g.,* so as to prepare to head home from work. Users can instruct the system to forego alerts if they are sensed to be present, working on a desktop system.

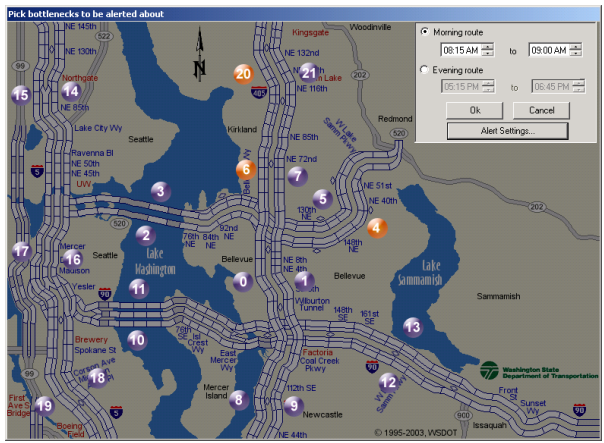

Figure 8: Desktop route-centric alert configuration tool, showing highlighting of an AM commuting route with associated active monitoring period.

As indicated in Figure 9, users can also sign up for alerts that indicate that an *unexpected situation* has arisen or will arise on their active route. We will now turn to our efforts to model and relay predictions about surprises.

## 3 Reasoning and Alerting about Surprises

Predictions about states of the world are most valuable to people when they complement a user's knowledge, rather than providing them with redundant information, given a user's intuitions and expectations. Many commuters in the Seattle area have an overall sense for the status of hotspots and the overall times until congestion at bottlenecks will likely start and end, based on their long-term experiences. To enhance the value of JamBayes for people who may be familiar with typical traffic patterns, we worked to include in JamBayes inferences about whether traffic states of interest would be viewed as *surprising*.

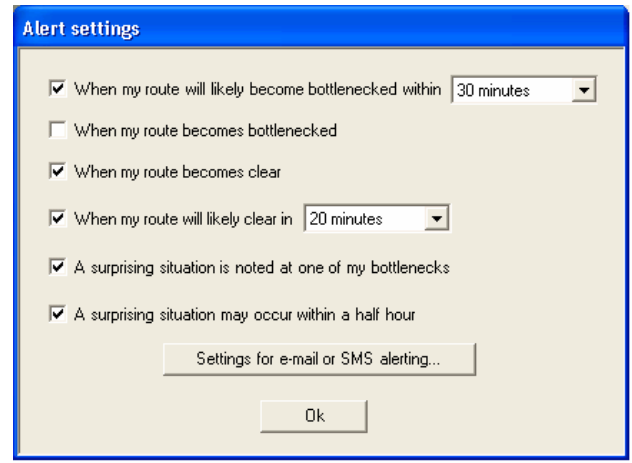

Figure 9: Policy configuration screen for mobile and desktop alerting.

There has been a great deal of work in detecting anomalies in data sets (*e.g.*, Bay and Pazzani, 1999, Wong, Moore, Cooper, *et al.*, 2003). Rather than seeking to identify anomalies in data, we are primarily interested in a model for states of the world that would likely surprise a typical user of Smartphlow. And moving beyond modeling when a user might be surprised by a current state of affairs, we have also been interested in models that can predict *future surprises*— even if everything appears to be quite normal now.

To identify situations that users would likely find surprising, we need to define a user model for surprise. In the current system, we use marginal models, that capture basic statistics for congestion, as an approximate model of user expectations. We employ as a user model the marginal statistics describing the status of each bottleneck for every 15 minute segment of time within each day of the week. The 15 minute marginal models are constructed from a small subset of observations that the machine-learning procedures use. Beyond time of day and day of week, we consider weather and holiday status. The latter pieces of evidence are widely available to commuters. To identify surprises, we compare the output of the marginal models with the real-time states to identify rare flows and congestion. We mark these situations as situations that would likely be surprising to users.

For surprise modeling in JamBayes, the current status of each bottleneck is compared to the likelihood of the observed state from the perspective of the marginal user model. If the likelihood of an observed open flow or congestion at a bottleneck occurs with a probability of 0.10 or less, we mark the situation as a surprising situation. We then use the surprise tag in alerting and display. When a surprising situation is noted, we overlay an exclamation point on the clock graphic displayed on the smartphone. Two examples of visualizations of surprise are displayed in Figure 10. On the left side of the figure, an exclamation point on a clock icon, predicting that a blockage will last 40 minutes, lets the user know this blockage is for this time in the early afternoon on a Sunday afternoon. On the right side of the figure, a surprise icon indicates that a portion of the highway that would be expected to be bottlenecked during rush hour is now wide open. If one of these regions had been part of a user's specified commute and the surprising situation had occurred during the hours configured for active route monitoring, the user would be alerted.

Before moving on, we note that there is opportunity to employ more sophisticated methods to develop and validate user models that could predict when congestion and flows would be surprising to different users or groups of users. For example, we could employ similar modeling methods to those described by Horvitz and Barry (1995) in a NASA Mission Control advisory system, named Vista. The work on Vista introduced a formal model of the value of *information revelation*. In that work, graphical models were constructed and refined to make inferences about the beliefs of users about the health of propulsion systems on the space shuttle. The user model for shuttle failures was used in conjunction with a detailed probabilistic base-level model of the propulsion system to dynamically configure a display of relevant information updates.

For the case of models of surprise for traffic, we might allow users in a future version of Smartphlow to specify a definition of surprise in terms of the rarity of an event, rather than using a default probability for anomalies. Additionally, we can extend the system to provide users with the ability to select from a set of user models, or even to customize a user model to capture additional details about his or her personal knowledge. For example, it is feasible to make available models more sophisticated than the marginal we have described (*i.e.*, one taking into consideration widely known observations about time of day, day of week, holidays, and weather). As an example, we could provide users who follow major sporting events and who have a sense for the basic relationships between traffic delays and these major events, with the ability to select a user model for surprises that uses a marginal model that includes sporting events.

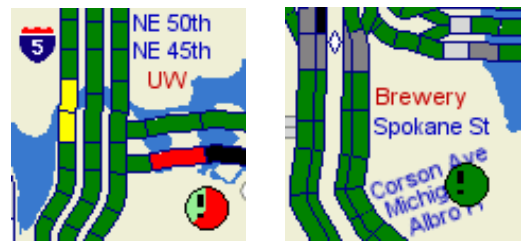

Figure 10: Display of situations identified as surprises.

## 4  Surprise Forecasting

We have reviewed a method for identifying current traffic situations that would likely surprise commuters, and reviewed how the current version of JamBayes, in use by a large population of users, can be instructed to render visual and audio alerts about specific bottlenecks or routes that users care about when such surprises arise. We now move to the realm of *future surprises*, and address the construction of models that can predict future surprise. We make such models of future surprise available for alerting users in the current version of the system.

### 4.1 Learning and Using Models of Future Surprises

To construct models of future surprises, we first identify sets of surprising situations, as described earlier. However, rather than simply use the information for alerting and then discard it, we store a case library of surprising events. We also consider observations available to the system at earlier points in time and add these to the cases. That is, the case library for future surprise contains sets of surprising events, coupled with observations that had been taken at different amounts of time before the occurrence of surprising outcomes. For example, we note all observations that are available to the system about traffic and its recent history at 30 minutes before a surprising event is noticed, per a definition of surprising event. The observations include the available stream of information about multiple sources of information, including the weather, incident reports, major events, and status of observations about the history of traffic flow throughout the Seattle traffic system. However, unlike the modeling efforts we have described in other sections, the evidence for the case is drawn from points in time at least 30 minutes before the surprising event.

Figure 11 schematizes the overall approach to learning to predict surprises in the future. Figure 12 displays a Bayesian network learned with this methodology for inferring the likelihood that states defined as surprising will occur in 30 minutes. We can use this model to explore the influence on the likelihood of surprises occurring 30 minutes in the future, given a report that a car accident just happened at bottleneck 15. The variable representing the accident at bottleneck 15 is highlighted at the center of the model in Figure 12.

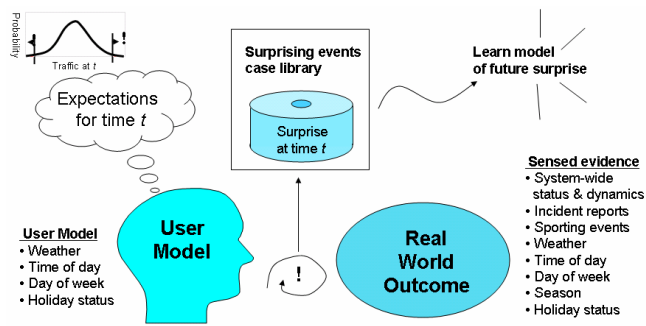

Figure 11: Learning models of future surprises. A user model defining surprising outcomes is used to compose a case library of surprises. These cases, along with detailed observations from the past are stored, and models of future surprise are constructed.

Exploring the predictions of the model reveals that an accident at bottleneck 15 influences the likelihood that there will be a surprising traffic situation at bottleneck 15 in 30 minutes—as might be expected by someone who observed the accident. However, the model also predicts that there is a significant likelihood of seeing surprising flows at bottlenecks 4, 11, and 17 at 30 minutes in the future.[1]

For a better view of the flows, Figure 13 shows the numbered bottlenecks overlayed on a map of the Seattle traffic system, indicating the site of the reported accident with a star (southbound on Hwy 5, north of Hwy 520). Bottlenecks predicted to have a significant increase in the likelihood of surprising flows in 30 minutes after the accident at bottleneck 15 are circled. The influences highlight the time-delayed propagation of effects through the Seattle traffic network. For example, probing the predictive model reveals that, for scenarios of an accident occurring at bottleneck 15 during several spans of time during the day, there will be a 0.5 probability of a surprising *low traffic flow* at bottleneck 4, across Lake Washington, 30 minutes after the accident.

### 4.2 Accuracy of Models of Future Surprise

To get a sense for the discriminatory power of the learned models of future surprise, we constructed a model of future surprise with seventy-five percent of the cases in the surprise case library and tested the ability to predict surprises 30 minutes into the future with the remaining twenty-five percent of the data.

---

[1] As demonstrated by the model, the accident at bottleneck 15 has a probabilistic relationship with the flow at bottleneck 18. We leave the explanation of this influence as an exercise to the reader.

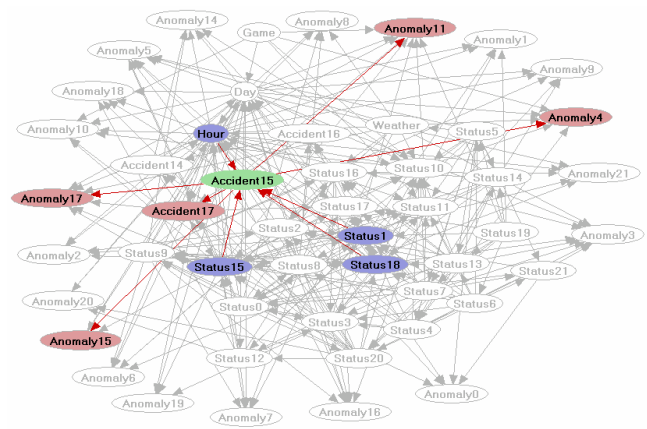

Figure 12: Predictive model for surprises 30 minutes into the future, showing influences of a current accident at bottleneck 15 on the likelihood of seeing surprising flows at several other bottlenecks 30 minutes later.

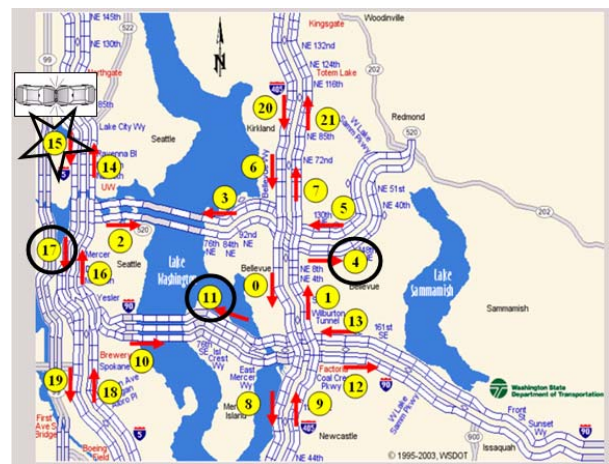

Figure 13: Display of bottlenecks, highlighting the influences of an accident at bottleneck 15 (star), on the likelihood of seeing a surprising traffic situation at the circled bottlenecks in 30 minutes, as described by the graphical model in Figure 12.

Classification accuracy does not provide a valuable signal as the accuracy is invariably reported as high for the marginal model (which assumes *no surprise*) given the rarity of surprises. Thus, we sought to visualize the relationship between the false negatives and false positives for surprises. Such curves for the task of predicting surprising traffic situations 30 minutes into the future for each bottleneck are displayed in Figure 14. We have removed from consideration all cases where the bottleneck was associated with a current surprising situation. Thus, the model shows relationships among false negatives and false positives for cases where a bottleneck is not showing anomalous flows at the current time.

We note from the graphs that, if we tolerate missing alerts about half of the surprises in 30 minutes, the false positive rate drops to approximately 0.05 for many of the bottlenecks. This means that the probabilistic models of future surprise can be harnessed to report about half of the surprises on the Seattle traffic system with a 0.05 false positive rate.

We integrated the future surprise reporting into the JamBayes service and, as highlighted in the options displayed in Figure 9, users can set up their systems to provide alerts when there will be a likely surprisingly light flow or surprising congestion in the future.

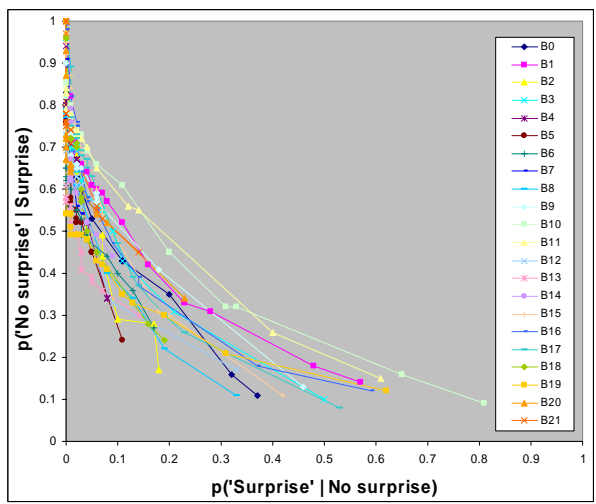

Figure 14: Surprise forecasting analysis. Curves show the relationship between false negatives ($y$ axis) and false positives ($x$ axis) for predictions of a surprise 30 minutes in the future for the 22 Seattle bottlenecks, for cases where no surprise currently exists at the bottlenecks.

## 4  Summary and Future Work

We described our efforts to construct and field a probabilistic traffic forecasting system named JamBayes. The system has been made available within our organization and is now in active use by over 2,500 people. We are excited to provide the fruits of learning and reasoning as a basic utility that provides ongoing daily guidance to a large number of people.

We are continuing to refine the methods. Our ongoing research includes investigating alternate machine learning modeling methods, such as exploring the value of boosting, and also considering extensions that explore other inference and modeling methodologies, including particle filtering, continuous time Bayesian networks (Nodelman, Shelton, and Koller, 2003, Nodelman and Horvitz, 2003), and queue-theoretic techniques. We are also pursuing a deeper understanding of models of future surprise. We are intrigued by the challenges and opportunities associated with building models of future surprises, for traffic congestion as well as in other domains.

**Acknowledgments**

We thank the WDOT for data and assistance. We thank Max Chickering for his assistance and enthusiastic support. We thank David Heckerman, Chris Meek, and Bo Thiesson for discussions and exploratory evaluation of boosting methods for traffic predictions.